\documentclass[11pt]{article}
\usepackage{coling2020}
\usepackage{times}
\usepackage{url}
\usepackage{latexsym}
\usepackage[textsize=tiny]{todonotes}
\usepackage{amsmath}
\usepackage{amsfonts}
\usepackage{booktabs}
\usepackage{multirow}
\usepackage{wrapfig}
\usepackage{subfigure}
\usepackage{caption}

\newcommand{\scene}{\ensuremath{\mathcal{S}}}
\newcommand{\categories}{\ensuremath{\mathcal{C}}}
\newcommand{\objects}{\ensuremath{\mathcal{O}}}
\newcommand{\objemb}{\ensuremath{\mathbf{v}}}
\newcommand{\catemb}{\ensuremath{\mathbf{c}}}
\newcommand{\emb}{\ensuremath{\mathbf{z}}}

\newcommand{\D}{\ensuremath{D_{\theta}}}
\newcommand{\En}{\ensuremath{E_{\phi}}}

\newcommand{\lossr}{\ensuremath{\mathcal{L}_{\mathsf{REC}}}}

\newcommand{\lossreg}{\ensuremath{\mathcal{L}_{\mathsf{REG}}}}

\newcommand{\lossimg}{\ensuremath{\mathcal{L}_{\mathsf{IMG}}}}
\newcommand{\guesswhat}{\textit{GuessWhat?!}}
\newcommand{\compguesswhat}{\textit{CompGuessWhat?!}}
\newcommand{\grolla}{\textsc{GroLLA}}

\colingfinalcopy

\title{Imagining Grounded Conceptual Representations from Perceptual Information in Situated Guessing Games}

\author{Alessandro Suglia$^1$, Antonio Vergari$^2$, Ioannis Konstas$^1$, Yonatan Bisk$^3$, \\\textbf{Emanuele Bastianelli}$^1$, \textbf{Andrea Vanzo}$^1$, \and \textbf{Oliver Lemon$^1$}\\
  $^1$Heriot-Watt University, Edinburgh, UK \\
  $^2$University of California, Los Angeles, USA \\
  $^3$Carnegie Mellon University, Pittsburgh, USA \\
  $^1$\texttt{\{as247,i.konstas,a.vanzo,e.bastianelli,o.lemon\}@hw.ac.uk} \\
  $^2$\texttt{aver@cs.ucla.edu},
  $^3$\texttt{ybisk@cs.cmu.edu} \\}

\begin{document}
\maketitle

\begin{abstract}
In visual guessing games, a Guesser has to identify a target object in a scene by asking questions to an Oracle. 
An effective strategy for the players is to learn conceptual representations of objects that are both discriminative and expressive enough to ask questions and guess correctly.
However, as shown by \newcite{suglia2020compguesswhat}, existing models fail to learn truly multi-modal representations, relying instead on gold category labels for objects in the scene both at training and inference time. 
This provides an unnatural performance advantage when categories at inference time match those at training time, and it causes models to fail in more realistic “zero-shot” scenarios where out-of-domain object categories are involved. To overcome this issue, we introduce a novel “imagination” module based on Regularized Auto-Encoders, that learns context-aware and category-aware latent embeddings without relying on category labels at inference time. 
Our imagination module outperforms state-of-the-art competitors by 8.26\% gameplay accuracy in the CompGuessWhat?!\ zero-shot scenario~\cite{suglia2020compguesswhat}, and it improves the Oracle and Guesser accuracy  by 2.08\% and 12.86\% in the \guesswhat{}\ benchmark, when no gold categories are available at inference time. The imagination module also  boosts reasoning about object properties and attributes.
\end{abstract}

\section{Introduction}
\label{intro}

\blfootnote{
    \hspace{-0.65cm}  
     This work is licensed under a Creative Commons 
     Attribution 4.0 International License.
     License details:
     \url{http://creativecommons.org/licenses/by/4.0/}.
}

Humans do not learn conceptual representations from language alone, 
but from a wide range of situational information~\cite{beinborn2018multimodal,bisk2020experience} as highlighted also by property-listing experiments~\cite{mcrae2005semantic}. 
When humans experience the concept of ``boat'', they \textit{simulate} a new representation by reactivating and aggregating \textit{multi-modal} representations that reside in their memory and are associated with the concept of ``boat'' (e.g., what a boat looks like, the action of sailing, etc)~\cite{barsalou2008grounded}. 
This simulation process is called \textit{perceptual simulation}.  
Therefore, it is no wonder that recent trends in learning conceptual representations adopt multi-modal and holistic approaches~\cite{bruni2014multimodal} wherein abstract distributional lexical representations~\cite{landauer1997solution,laurence1999concepts} learned from text corpora 
are augmented or refined with \textit{perceptual information} for concrete and context-aware representations built from visual~\cite{kiela2018learning,lazaridou2015combining}, olfactory~\cite{kiela2015grounding}, or auditory~\cite{kiela2015multi} modalities. 

Language games between AI agents, inspired by Wittgenstein's Language Games among humans~\cite{wittgenstein1953philosophische},  are an excellent  test bed for such approaches since concepts are expected to emerge when agents are required to communicate to solve specific tasks in specific environments.
\guesswhat{}~\cite{de2017guesswhat} is a prototypical language game of this kind: a Guesser has to identify a target object in a scene represented as an image by asking questions to an Oracle. Learning to ground pixels of the scene into object representations that are relevant for the object category they belong to (\textit{category-aware}), but are also particularized for the specific scene (\textit{context-aware}), is fundamental for the Guesser to effectively converse with the Oracle and vice-versa.

\begin{figure}
\centering
\includegraphics[keepaspectratio, scale=0.8, trim={0 20pt 0 0pt}, clip]{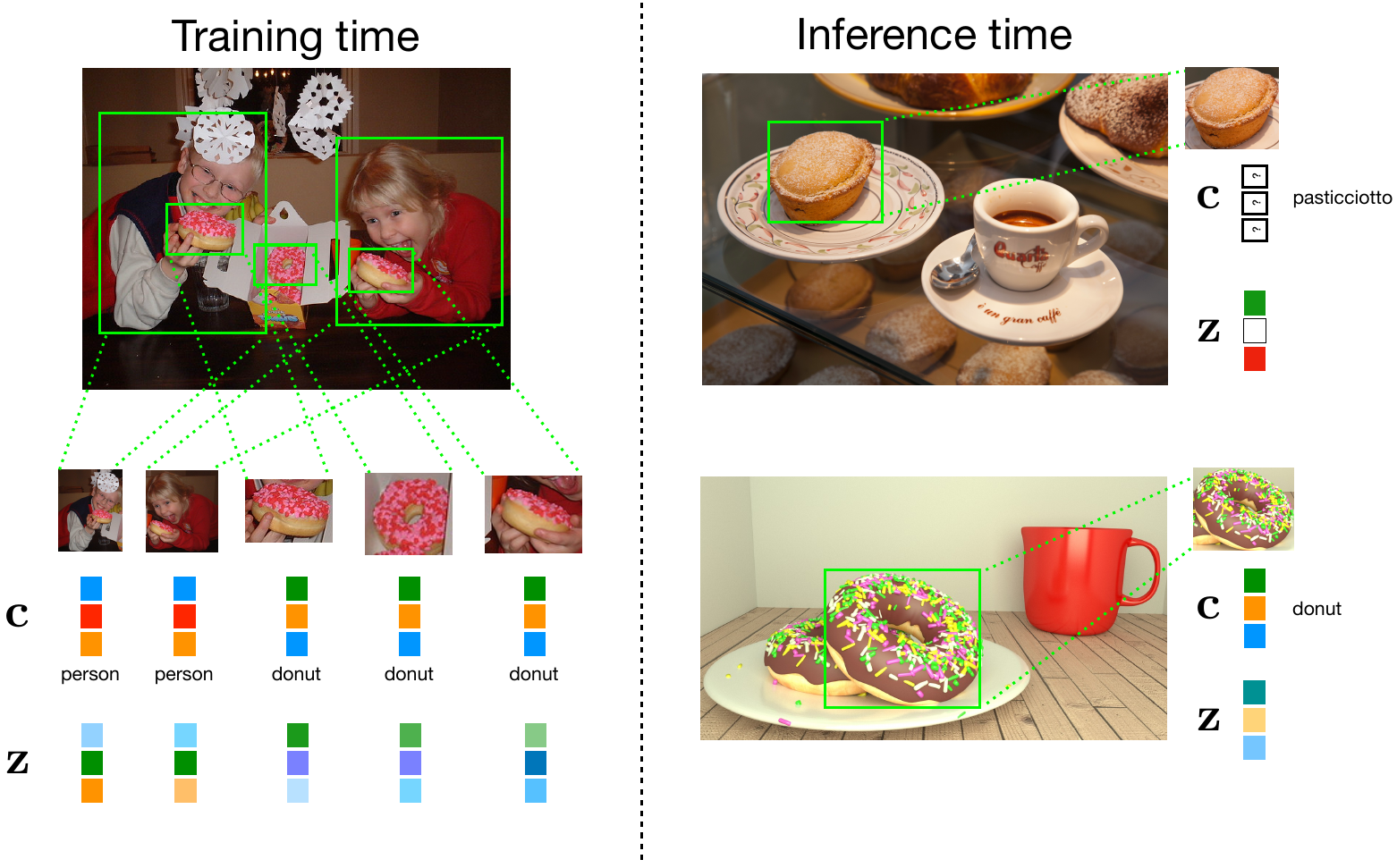}
\caption{Common approaches to visual grounding such as  \newcite{de2017guesswhat} and \newcite{zhuang2018parallel} rely on gold category labels at test time, thereby failing to ground novel objects from categories not seen during training (e.g., a ``pasticciotto'', top right) or to properly encode known categories but with unseen visual features (like a ``frosted donut'', bottom right) since they employ \textit{category embeddings} $\catemb$ from a predefined set that are fixed for each object.
Instead, embeddings $\emb$ learned by our imagination module can be flexibly category-aware allowing them to generalize to unseen categories. 
}
\label{fig:imagination_overview}
\end{figure}

We consider a model truly \textit{multi-modal} if it always uses all the modalities to make decisions. However, existing approaches \cite{de2017guesswhat,shekhar2019beyond} rely instead on \textit{gold category labels} that are assumed to be available also at inference time, thus making these models depend on this modality and discarding the others.
This not only poses an unnatural performance advantage for players in controlled benchmark scenarios like the \guesswhat{}~game when categories at inference time match those at training time, but causes them to fail in more realistic zero-shot scenarios~\cite{suglia2020compguesswhat} where players are required to generalize to out-of-domain object categories. For example, consider an agent that during training has only seen glazed donuts, associated with the fixed ``donut'' category embedding (cf. Figure~\ref{fig:imagination_overview}).
At inference time, the model cannot ground visual representations for objects belonging to the ``pasticciotto'' (an Italian pastry) category, since such a category was not in its repertoire.
Similarly, it will likely represent frosted donuts with a generic ``donut'' embedding, despite the perceptual differences among different types of donut.

In this paper, we tackle the above limitations by introducing a novel \textit{imagination module} based on Regularized Auto-encoders~\cite{ghosh2019variational}, which are able to derive \textit{imagination embeddings} directly from perceptual information in the form of the object crop. Our formulation of the reconstruction loss allows the model to learn \textit{context-aware} and \textit{category-aware} imagination embeddings.
Thus, removing the need for gold category labels at inference time and greatly improving zero-shot generalization.
Section \ref{sec:compguesswhat_evaluation} integrates our imagination component into the Oracle model of \newcite{de2017guesswhat} and the Guesser model of \newcite{shekhar2019beyond}.
We show that the new imagination models are state-of-the-art in the recently introduced \compguesswhat{}\ benchmark~\cite{suglia2020compguesswhat} outperforming current models by $8.26\%$.
It also improves the Oracle's  and Guesser's accuracy (by $2.08\%$ and $12.86\%$, respectively) in the standard \guesswhat{} when no gold category labels are available. Lastly, we show that imagining latent object representations greatly helps to reason about object visual properties (i.e., color, shape, etc.), qualifying our module as a generic \textit{perceptual simulation} component al\`a \newcite{barsalou2008grounded}.

\section{Background: Guessing Games and Concept Representations}
\label{sec:background}

\guesswhat{}\  is an instance of a multi-word guessing game~\cite{steels2015talking}. 
Every game involves two players: an \textit{Oracle} and a \textit{Guesser} conversing about a \textit{scene} $\scene$ (a natural image).
A {scene} $\scene$ can be abstracted into a collection of \textit{objects} $\objects$, 
each of which is associated with a category $c_{i}\in\categories, i=\{1,\ldots,K\}$.
The aim of the Guesser is to identify a \textit{target object} $o^{\mathbf{*}} \in \objects$ by asking questions about $\scene$ to the Oracle. 
The gameplay of \guesswhat{}\   thus comprises three tasks: i) \textit{question generation} where the Guesser inquires about an object in the scene $\scene$ given the dialogue generated so far; ii) \textit{answer prediction}, where the Oracle answers $a \in \mathcal{A} = \{\mathsf{Yes}, \mathsf{No}, \mathsf{N/A}\}$ given the scene $\scene$, question and the target object $o^{\mathbf{*}}$;  and iii) \textit{target prediction} where 
the Guesser selects a candidate object with the highest relevance score $r(o_i)$.

Several architectural variants have been proposed to tackle \guesswhat{}\  (cf. Section~\ref{sec:related} for some related works).
In this work we adopt the recent GDSE model~\cite{shekhar2019beyond}, which learns a
visually grounded dialogue state used to learn both question generation and target object prediction. As shown below, GDSE does not deliver the desired multi-modality needed, therefore we extend it with our Imagination component to obtain more effective multi-modal object representations.

For successful gameplay, both the Guesser and Oracle must build representations of the scene 
that contain 
specific perceptual information of objects (object-aware),
are relevant for the object category they belong to (category-aware), and are specialized to the scene in which the game is played (context-aware).
As the scene $\scene$ is an image, it is natural to associate each object $o_{i}\in\objects$ with a \textit{perceptual embedding}, i.e., a vector $\objemb_{i}\in\mathbb{R}^{d_{\objects}}$ extracted from the penultimate layer of a pretrained vision model (e.g. 
ResNet-152 ~\cite{shekhar2019beyond}) based on their bounding box.\footnote{
Bounding boxes are assumed to be given, e.g. by using object recognition as a pre-processing step \cite{anderson2018bottom}.}

However, these representations are not sufficient as they are neither \textit{context-aware} nor \textit{category-aware}, i.e., they ignore other objects in the scene and do not leverage their category information. %
GDSE and other recent approaches~\cite{de2017guesswhat,shekhar2019beyond,zhuang2018parallel,shukla2019should} coped with the second issue by introducing \textit{category embeddings} as $d_{\categories}$-dimensional continuous representations $\catemb_{k}\in\mathbb{R}^{d_{\categories}}$ for $k=1,\ldots,K$.
Once learned, a category embedding $\mathbf{c}$ is then concatenated to an 8-dimensional feature vector $\mathbf{s}_i$ derived from the object bounding box (cf.~\newcite{de2017guesswhat}).
While these embeddings partially solve category-awareness, they are \textit{not object-aware}.
For instance, the embedding for the object category ``apple'' will be the same regardless of a particular object to be a red or green apple, i.e., most likely a centroid representation of the objects seen only during training.
Moreover, if during training we only see red apples, at inference time, we will likely fail to detect green apples as belonging to the same category (Figure~\ref{fig:imagination_detail}).
These issues have gone unnoticed since category embeddings usually boost performances on the original \guesswhat{}\ task, given that gold category labels are also available at inference time.
However, this boost is illusory: models relying on this symbolic information to be always available are not learning to exploit all modalities.
In fact, a $~20\%$ drop in the Guesser accuracy if gold category labels are not provided has been reported in~\newcite{zhuang2018parallel} for~\guesswhat{}\ and analogous poor results in more 
realistic benchmarks measuring zero-shot generalization such as \compguesswhat{}~\cite{suglia2020compguesswhat}.

\section{Imagination Module: Learning Context- and Category-aware Object Representations}
\label{sec:imagination}
\begin{figure}
	\centering
	\subfigure[Imagination module]{
		\includegraphics[width=.7\linewidth]{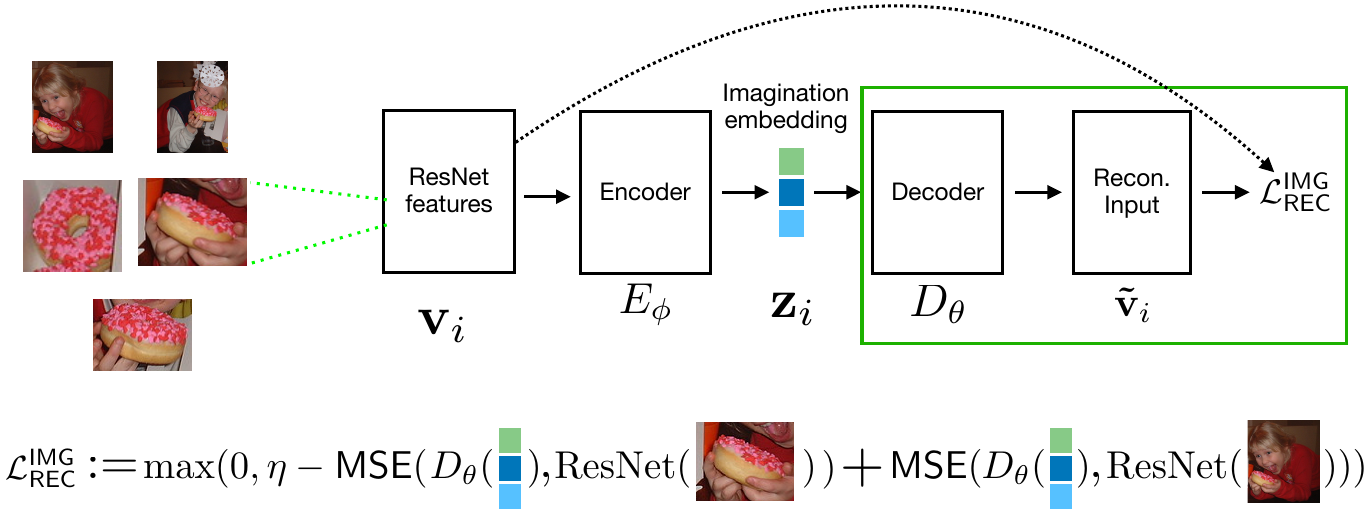}
		\label{fig:imagination_detail}
	}
	\subfigure[Oracle model]{%
		\includegraphics[width=0.35\linewidth]{img/imagination_oracle.pdf}
		\label{fig:oracle_model}
	}
	\subfigure[Guesser model]{
		\includegraphics[width=0.5\linewidth]{img/imagination_guesser.pdf}
		\label{fig:guesser_model}
	}
	\caption{Imagination-based Representation Learning: Given the perceptual information  $\mathbf{v}_i$ of object $o_i$, we learn an imagination embedding $\emb_i$ generated by Encoder $\En$. The latent code is optimized to reconstruct the original visual representation $\mathbf{v}_i$ (the ``donut" ResNet encoding) via the reconstruction loss $\lossr^{\mathsf{IMG}}$ using the Decoder $\D$. Figures~\ref{fig:oracle_model} and~\ref{fig:guesser_model} show how the imagination embedding $\emb$ replaces the category embedding $\catemb$ in the Oracle model from \newcite{de2017guesswhat} and Guesser model from \newcite{shekhar2019beyond} respectively, and is concatenated to the spatial information $\mathbf{s}_i$.}
	\label{fig:imagination}
\end{figure}

To overcome the limitations of GDSE and competitors
and realize a form of \textit{perceptual simulation} in a learning system, we introduce a generic component---named the \textit{imagination module}---which learns latent concept representations that are both context- and category-aware, without relying on category labels at inference time.
Our imagination model can be understood in the context of representation learning via deep generative models~\cite{bengio2013representation} which has been popularized by variational autoencoders (VAEs) \cite{kingma2013auto,kingma2014semi}, and GANs~\cite{goodfellow2014generative}.
Specifically, we substantially extend the recently introduced regularized autoencoders (RAEs) framework~\cite{ghosh2019variational}.
%
RAEs are simplified VAEs where stochasticity in the encoder and decoder is dropped in favor of more stable training and more informative embedding learning.
In fact, RAEs do not suffer from several issues known to affect VAEs, such as poor convergence and the possibility of learning embeddings that are independent of the input images (cf.~\newcite{ghosh2019variational} for a detailed discussion).
More crucially 
for our purposes, RAEs do not have to compromise the informativeness of the learned embeddings with a fixed a-priori structure in the latent space that enables simple sampling (e.g.,  an isotropic Gaussian prior).
VAEs which need to have such a fixed prior, instead, are deemed to learn embeddings that are less informative w.r.t. objects, categories, and context information.

\paragraph{Module architecture.}
Figure~\ref{fig:imagination_detail} summarizes our imagination module.
Its aim is to distill a context and category-aware 
embedding $\emb_{i}\in\mathbb{R}^{d_{Z}}$ per object $o_{i}$ in scene $\scene$. 
To this end, we adopt an encoder $\En$ parameterized by $\phi$ that maps a perceptual embedding $\objemb_{i}$ of object $o_{i}$ to its imagined counterpart $\emb_{i}$, i.e., $\En(\objemb_{i})=\emb_{i}$. A decoder $\D$ realizes the inverse mapping $\tilde{\objemb}_{i}=\D(\emb_{i})$, with $\tilde{\objemb}_{i}\in\mathbb{R}^{d_{\objects}}$ being also called the reconstruction of the input $\objemb_{i}$.
As in RAEs, our per-object loss $\lossimg$ comprises a reconstruction loss ($\lossr$), weighting how good the reconstructions of $\D$ are w.r.t. the encoded representations by $\En$, and a regularization term ($\lossreg$) enhancing generalization by smoothing the decoder $\D$.
This leads to the following composite loss:
\begin{equation}
    \lossimg = \lossr + \alpha\lossreg,
    \label{eq:loss}
\end{equation}
where $\alpha$ is an hyperparameter controlling regularization.\footnote{
\newcite{ghosh2019variational} use two different hyperparameters for the two terms in $\lossreg$ Optimizing them independently had no evident benefit in our experiments, hence we simply treat them as a single regularizer together.
}
As in $L_{2}$-RAE~\cite{ghosh2019variational}, the regularization component is defined as $\lossreg:=||\emb_{i}|| + ||\theta||_{2}$: the first term bounds the latent embedding space learned by $\En$ easing optimization; the second enforces smoothing over $\D$ improving generalization over regions of the latent space that are unseen during training.

Differently from RAEs, we devise a specific reconstruction loss tailored to learn contextual and category-aware representations.
In conventional RAEs, in fact, the reconstruction loss is defined as the Mean Squared Error (MSE) representing the distance between $\objemb_{i}$ and its reconstruction $\tilde{\objemb}_{i}$, so that $\lossr^{\mathsf{RAE}}:= \mathsf{MSE}(\objemb_i, \tilde{\objemb}_{i})$.
This loss is purely unsupervised and as such agnostic to object categories or to the scene context.
To our aims, we define a custom \textit{imagination reconstruction loss} $\lossr^{\mathsf{IMG}}$ as an instance of a max-margin triplet-loss~\cite{wang2014learning,schroff2015facenet}, as follows. 
Let $c_{i}$ be the category of object $o_{i}$ with perceptual embedding $\objemb_{i}$ in scene $\scene$ and let $\objects_{\neg c_{i}}=\{o_{j}\mid o_{j}\in\objects \wedge c_{j}\neq c_{i}\}$ be the set of all objects in $\scene$ belonging to a different category than $c_{i}$.
Our per-object $\lossr^{\mathsf{IMG}}$ term is defined as:
\begin{align}
    \lossr^{\mathsf{IMG}} &:= \max(0, \eta - \mathsf{MSE}(\objemb_i, \D(\emb_i)) + \mathsf{MSE}(\objemb_{j}, \D(\emb_i))),
\end{align}
where $\eta$ is the minimum margin between two components: \textit{i}) the distance between the perceptual embedding $\objemb_{i}$ and its reconstruction $\D(\emb_i)$, and \textit{ii}) the distance between the perceptual embedding $\objemb_{j}$ of a randomly sampled object $o_{j} \in \objects_{\neg c_{i}}$ and the reconstruction $\D(\emb_i)$.
By doing so, we enforce each object representation to be \textit{representative of its category given a specific context} by locally contrasting it to another object of a different category in the same scene.
Note that this is strikingly different from previous approaches employing a max-margin loss~\cite{elliott2017imagination,kiros2018illustrative} where ``negative'' objects are arbitrarily sampled from other scenes in the same batch.

\paragraph{Imagining at inference time.} Differently from the category embeddings $\mathbf{c}$ employed by all previous work, our imagination embeddings $\emb$ \textit{do not depend on gold category labels at inference time}, while still being context-aware and category-aware.
In fact, once parameters $\phi$ have been learned, the encoder $\En$ contains all the information needed to distill embeddings $\emb$ independently of $\lossimg$, which is necessary \textit{only} at training time. We consider \textit{imagination} the ability of the model of generating latent representations on-the-fly. 
Therefore, for both Guesser and Oracle models we consider an object representation for object $o_{i}$ that replaces $\mathbf{c}_{i}$ with $\emb_{i}$ and concatenates it with its spatial information $\mathbf{s}_i$ (see Figures \ref{fig:oracle_model} and \ref{fig:guesser_model} and Appendix~\ref{appendix:model_details} for details).
By doing so, we consider every gameplay situated in a reference scene as an experience where our imagination module is able to derive a latent conceptual representation simply by ``looking" at objects, realizing a \textit{perceptual simulator}~\cite{barsalou2008grounded}. 
We plan to investigate how to combine label-dependent category embeddings $\mathbf{c}$ with our imagination embeddings $\emb$, similarly to how some VAE variants tackle semi-supervised classification scenarios~\cite{kingma2014semi}.

\section{Experimental Investigation}\label{sec:evaluation}

To assess the impact of using the imagination embeddings against the category embeddings, we use two evaluation benchmarks: \guesswhat{}\ and \compguesswhat{}. More information about the training procedure can be found in Appendix~\ref{appendix:training_details}.

\subsection{\guesswhat{} Evaluation}\label{sec:guesswhat_evaluation}
In this experiment, we evaluate the accuracy of the Oracle in answering questions and the accuracy of the Guesser in selecting the target object.
We consider as both training and evaluation data all the gold dialogues (and questions) that have been labeled as successful in the dataset~\cite{de2017guesswhat}. We want to highlight that in this evaluation phase, the models using label-aware object encodings have gold information both at training and test time. This is true both for the Oracle and Guesser models. However, this does not hold for all other models using the imagination component. 

\subsubsection{Experimental Setup}
\paragraph{Oracle task.}

We evaluate the imagination-based Oracle and compare it to several combinations of the 
following baselines with and without category embeddings 
from \newcite{de2017guesswhat}: 
1) \textsc{Majority}: majority classifier; 2) \textsc{Question}: uses only the question; 3) \textsc{Image}: uses only the image representation; 4) \textsc{Crop}: uses only the crop representation of the target object. 

\paragraph{Guesser task.}
Similarly, we compare the GDSE model using imagination embeddings (\textsc{GDSE+imagination}) with the following \textit{label-aware} baselines:
1) text-only baselines using LSTM encoder (\textsc{LSTM}) and Hierarchical Recurrent Encoder-Decoder architecture~\cite{serban2017hierarchical} (\textsc{HRED}) as well as their corresponding multi-modal models \textsc{LSTM+Image} and \textsc{HRED+Image}; 2) \textsc{ParallelAttention}~\cite{zhuang2018parallel} and \textsc{GDSE}~\cite{shekhar2019beyond}. We also compare with variants of the above that do not use any category embeddings or gold category labels (\textsc{*-nocat}), as well as models with predicted category labels (\textsc{*-predcat}).\footnote{We train an object classifier using as input the ResNet-101 features generated for the object crop. It achieves $65\%$ accuracy evaluated on \emph{all} objects in the \guesswhat{}\ test set.}

\clearpage
\subsubsection{Results}\label{sec:oracle_results}

\begin{wraptable}[21]{r}{0.5\textwidth}
\vspace{-10pt}
\small
\centering
\textsc{
\begin{tabular}{@{}l@{\hspace{3pt}}l@{\hspace{6pt}}cc@{}}
& \textbf{Model}  & \textbf{Val} & \textbf{Test} \\ 
\toprule
\multicolumn{1}{@{}l@{\hspace{3pt}}}{\multirow{4}{*}{\rotatebox{90}{\textbf{BASE}}}}   & Majority                              & 53.80\%                      & 49.10\%                \\
\multicolumn{1}{l}{}                                      & Ques                              & 58.30\%                      & 58.80\%                \\
\multicolumn{1}{l}{}                                      & Img                                 & 53.30\%                      & 53.30\%                \\
\multicolumn{1}{l}{}                                      & Crop                                  & 57.30\%                      & 57.00\%                   \\ 
\midrule
\multicolumn{1}{@{}l@{\hspace{3pt}}}{\multirow{5}{*}{\rotatebox{90}{\textbf{W/ CAT}}}}    & DV-Ques+Cat                   & 74.20\%                      & 74.30\%                \\ 
\multicolumn{1}{l}{}                                      & DV-Ques+Crop+Cat            & 75.60\%                      & 75.30\%                \\
\multicolumn{1}{l}{}                                      & DV-Ques+Spatial+Cat         & \textbf{78.90}\%                      & \textbf{78.50}\%                \\
\multicolumn{1}{l}{}                                      & DV-Ques+Spatial+Crop+Cat  & 78.30\%                      & 77.90\%                \\
\multicolumn{1}{l}{}                                      & DV-Ques+Spatial+Img+Cat & 76.80\%                      & 76.50\%                \\
\midrule \midrule
\multicolumn{1}{@{}l}{\multirow{6}{*}{\rotatebox{90}{{\small\textbf{MM}}}}} & DV-Ques+Crop                       & 70.90\%                      & 70.80\%                \\
\multicolumn{1}{l}{}                                      & DV-Ques+Img                      & 59.80\%                      & 60.20\%                \\
\multicolumn{1}{l}{}                                      & DV-Ques+Spatial                    & 68.80\%                      & 68.70\%                \\
\multicolumn{1}{l}{}                                      & DV-Ques+Spatial+Crop             & 74.00\%                         & 73.80\%                \\
\multicolumn{1}{l}{}                                      & DV-Ques+Spatial+Crop+Img     & 72.30\%                      & 72.10\%                \\
\multicolumn{1}{l}{}                                      & Imagination                 & \textbf{75.78}\%                      & \textbf{75.88}\%                \\ 
\bottomrule
\end{tabular}
}
\caption{Oracle results on gold questions: we compare the \textsc{Imagination} Oracle model to models from \newcite{de2017guesswhat} (\textsc{DV-*}). We group them into models relying on gold category labels (\textbf{W/ CAT}) and models that only use multi-modal perceptual information (\textbf{MM}).} 
\label{tab:oracle_gold_questions}
\end{wraptable}

\paragraph{Oracle task.}
In Table~\ref{tab:oracle_gold_questions}, we divide configurations into \textit{category-aware}~\cite{de2017guesswhat} and \textit{multi-modal}. The model reference for several other publications on \guesswhat{}\ is a \textit{category-aware} model \textsc{Question+Spatial+Category}. 
However, by relying on symbolic information in the form of category labels, it is inevitably not truly multi-modal anymore because the heavy-lifting is done by these embeddings. 
As shown in the results, other multi-modal models such as \textsc{Question+Spatial+Crop} and \textsc{Question+Crop}, are not able to learn effective representations to bridge the gap between category-aware and category-free models. On the other hand, the proposed imagination model is able to reduce this gap without relying on gold information as input. Indeed, we are able to learn category-aware and context-aware latent codes by using category information only in our loss function. 

We investigate this argument further by using a rule-based question classifier~\cite{shekhar2019beyond} to partition the test questions according to their type. Table~\ref{tab:oracle_per_question_type} summarizes this analysis; we include models considered truly multi-modal and the best Oracle model \textsc{Question+Spatial+Category}. The latter can answer with high accuracy questions about specific object instances (e.g., ``is it the dog?") or super-categories (e.g., 
``is it an animal?") since it is using category embeddings as input. 
However, when it comes to answering questions about perceptual properties of the target object, it loses some accuracy points because the perceptual information is missing from the category embedding representing a centroid of typical instances seen at training time only. On the other hand, the \textsc{Imagination} model is able to bring improvements of $1.34\%$, $5.81\%$, and $2.52\%$ for location, color, and shape questions, respectively. On questions related to perceptual information, models using crop information seem to be on par with the \textsc{Imagination} model. However, our model is able to obtain an improvement over \textsc{+CROP} of $1.84\%$ in object questions and of $1.11\%$ on super category questions solely by relying on the imagination embeddings. 

\begin{table}[]
\small
\center
\textsc{
\begin{tabular}{@{}l@{\hspace{2pt}}c@{\hspace{7pt}}c@{\hspace{7pt}}c@{\hspace{7pt}}c@{\hspace{7pt}}c@{\hspace{7pt}}|c@{\hspace{7pt}}c@{}}
& \multicolumn{5}{c}{\textbf{Perceptual Information}} & \multicolumn{2}{|c}{\textbf{Categorical Information}} \\\hline
\multirow{2}{*}{\textbf{Model} \begin{tiny}(DV-Ques+Spatial)\end{tiny}}  & 
\multirow{2}{*}{\textbf{Location}} & 
\multirow{2}{*}{\textbf{Shape}}  & 
\multirow{2}{*}{\textbf{Color}}   & 
\multirow{2}{*}{\textbf{Texture}} & 
\multirow{2}{*}{\textbf{Size}}    & 
\textbf{Super} & 
\multirow{2}{*}{\textbf{Object}}   \\
& & &  & & & \textbf{Category}   &   \\
\toprule
+ Crop            & 66.86\%  &	69.08\%	& 67.25\%	& 68.30\%	& \textbf{65.09\%}	& 88.94\%	& 80.48\% \\
+ Category        & 67.48\%           & 68.42\% & 61.83\%          & \textbf{70.08\%} & 60.14\%        &    \textbf{97.09\%}       & \textbf{88.82\%}          \\
+ Category + Crop  & 65.27\% &	60.34\%	& 59.14\% &	65.76\%	& 59.08\%	& 96.19\% &	86.32\%         \\
+ Imagination     & \textbf{68.62\%}  & \textbf{69.08\%}         & \textbf{67.64\%} & 69.86\%          & 62.65\%                 & 90.05\%          & 82.32\% \\ \bottomrule
\end{tabular}
}
\caption{Oracle accuracy grouped by question type for the best Oracle model with category information (\textsc{DV-Ques+Spatial}) and for multi-modal variants using either perceptual or categorical information.}
\label{tab:oracle_per_question_type}
\end{table}

\paragraph{Guesser task.}
Table~\ref{tab:guesser_gold_dialogues} compares several category-aware and multi-modal models;  \textsc{ParallelAttention} and \textsc{GDSE-SL} are the two best performing configurations. 
However, when \textsc{ParallelAttention} does not have access to category information (\textsc{ParallelAttention-nocat}) its performance drops by $3.7\%$ (also noted by \newcite{zhuang2018parallel}). 
We confirmed the same behavior for GDSE-SL as well (\textsc{GDSE-SL-nocat}),  noticing a more significant drop in performance of $16.95\%$ which is in line with the simpler \textsc{LSTM+Image} model. On the other hand, GDSE-SL with our imagination component (\textsc{GDSE-SL+Imagination}), performs comparably with the category-aware model and better then all multi-modal models.
Therefore we argue that it is possible to learn object representations that, given a representation for the current dialogue state, allow for discriminating the target object among other candidates \emph{without} relying on symbolic information.

\begin{wraptable}[16]{r}{0.5\textwidth}
\vspace{-35pt}
\small
\centering
\textsc{
\begin{tabular}{@{}llcc@{}}
& \textbf{Model}          & \textbf{Val} & \textbf{Test }       \\
\toprule
& Human                   & 90.80\%                      & \multicolumn{1}{c}{90.80\%} \\
\multicolumn{1}{l}{}                                      & Random                  & 17.10\%                      & \multicolumn{1}{c}{17.10\%} \\ \midrule
\multicolumn{1}{@{}l}{\multirow{7}{*}{\rotatebox{90}{\textbf{CATEGORY}}}}    & LSTM                    & 62.10\%                      & 61.30\%                      \\
\multicolumn{1}{l}{}                                      & HRED                    & 61.80\%                      & 61.00\%                         \\
\multicolumn{1}{l}{}                                      & LSTM+Image              & 61.50\%                      & 60.50\%                      \\
\multicolumn{1}{l}{}                                      & HRED+Image              & 61.60\%                      & 60.40\%                      \\
\multicolumn{1}{l}{}                                      & ParallelAttention       & 63.80\%                      & \textbf{63.40\%}    \\
\multicolumn{1}{l}{}                                      & GDSE-SL                 & 63.14\%                      & 62.96\%                     \\
\multicolumn{1}{l}{}                                      & GDSE-SL-predcat         & 52.08\%                      & 51.00\%                      \\
\midrule \midrule
\multicolumn{1}{@{}l}{\multirow{4}{*}{\rotatebox{90}{\textbf{MM}}}} & LSTM+Image-nocat        & 50.10\%                      & 48.60\%                      \\ 
 & ParallelAttention-nocat & 55.70\%                      & \textbf{59.70}\%                      \\
\multicolumn{1}{l}{}                                      & GDSE-SL-nocat           & 46.11\%                      & 46.01\%                      \\
\multicolumn{1}{l}{}                                      & GDSE-SL-imagination     & \textbf{59.54\%}             & 58.90\%             \\
\bottomrule
\end{tabular}
}
\caption{Guesser accuracy on successful gold dialogues: we compare \textsc{GDSE-SL-imagination} with i) models that are truly multi-modal (\textbf{MM}) and ii) use category information (\textbf{CATEGORY}).} 
\label{tab:guesser_gold_dialogues}

\end{wraptable}

\subsection{\compguesswhat{} Evaluation}\label{sec:compguesswhat_evaluation}
\compguesswhat{}\ is a benchmark proposed to assess the quality of models' representations and out-of-domain generalization. It includes the following tasks: a) \textit{in-domain gameplay accuracy}, -- selecting the target object with model generated dialogues as input, b) \textit{attribute prediction task} -- assessing the ability of the dialogue representation to recover target object attributes, and c) \textit{zero-shot gameplay accuracy} -- selecting the target object among objects belonging to categories never seen by the model during training. In contrast to \guesswhat{}, the attribute prediction and zero-shot tasks give us more insights about the quality of the learned representations and the model's generalization ability.

\subsubsection{Experimental Setup}
We compare imagination-based models with baselines used in \newcite{suglia2020compguesswhat}: 
1) \textsc{Random}: randomly selects an object; 2) \textsc{DeVries-SL}: presented in \newcite{de2017guesswhat} trained using Supervised Learning; 3) \textsc{DeVries-RL}: \textsc{DeVries-SL} with Questioner fine-tuned using Reinforcement Learning \cite{strub2017end}; and where  4) \textsc{GDSE-SL} and 5) \textsc{GDSE-CL} are the same as used in Section~\ref{sec:guesswhat_evaluation}. 

\subsubsection{Results}
\paragraph{In-domain gameplay.}
Table~\ref{tab:compguesswhat} presents the results on the \compguesswhat{}\ benchmark. Models are tasked to play the game by generating up to 10 questions and corresponding answers. Firstly, we note that the results for \textsc{GDSE-CL+Imagination}---the collaborative version of the model with Imagination---is still in the same ballpark of more complex models, such as \textsc{DeVries-RL} that is using category embeddings as input. At the same time, we notice that overall both imagination models perform worse than the \textsc{GDSE-*} models. We impute this drop to the introduction of additional loss terms that probably have changed the training dynamic of a cumbersome modulo-$n$ multi-task training~\cite{shekhar2019beyond}. This downside calls for a more principled way of handling tasks of different complexity (i.e., question generation and target prediction) in a multi-task learning system; we leave this for future work.   

\begin{table*}[t]
\small
\centering
\textsc{
\begin{tabular}{l@{}cccccccc}
\toprule
{} & \textit{{Gameplay}} & \multicolumn{4}{c}{\textit{{Attribute Prediction}}} & \multicolumn{2}{c}{\textit{{Zero-shot Gameplay}}} & \\
{} & \textbf{Accuracy} & \textbf{A-F1} & \textbf{S-F1} & \textbf{AS-F1} & \textbf{L-F1} & \textbf{ND-Acc} & \textbf{OD-Acc} & \multicolumn{1}{l}{{\textbf{GroLLA}}} \\   
\cmidrule(lr){2-2} \cmidrule(lr){3-6} \cmidrule(lr){7-8} \cmidrule(lr){9-9}
Random & 15.81\% & 15.1 & 0.1 & 7.8 & 2.8 & 16.9\% & 18.6\% & 13.3 \\
\midrule
DeVries-SL & 41.5\% & 46.8 & 39.1 & 48.5 & 42.7 & 31.3\% & 28.4\% & 38.5\\
DeVries-RL & 53.5\% & 45.2 & 38.9 & 47.2 & 42.5 & 43.9\% & 38.7\% & 46.2 \\
\midrule
GDSE-SL & 49.1\% & \textbf{59.9} & 47.6 & \textbf{60.1} & 48.3 & 29.8\% & 22.3\% & 43.0 \\
GDSE-CL & \textbf{59.8}\% & 59.5 & 47.6 & 59.8 & 48.1 & 43.4\% & 29.8\% & 50.1 \\
\midrule
GDSE-SL+Imagination & 43.82\% & 56.23 & 47.37 & 57.2 & \textbf{51.73} & 39.19\% & 39.90\% & 45.50 \\
GDSE-CL+Imagination & 51.98\% & 57.59 & \textit{47.6} & 58.31 & 50.42 & \textbf{46.56}\% & \textbf{46.96}\% & \textbf{50.74}\\
\bottomrule
\end{tabular}
}
\caption{Results for the \compguesswhat{} benchmark~\cite{suglia2020compguesswhat}. We assess model quality in terms of \textit{gameplay} accuracy, \textit{attribute prediction} quality, measured in terms of F1 for the \textit{abstract}~(\mbox{A-F1}), \textit{situated}~(S-F1), \textit{abstract+situated}~(AS-F1) 
and \textit{location}~(L-F1) prediction scenario, as well as \textit{zero-shot learning gameplay}. \grolla{} is a macro-average of the individual scores.}
\label{tab:compguesswhat}
\end{table*}

\paragraph{Attribute prediction.}
Table~\ref{tab:compguesswhat} reports the attribute prediction task results. In this scenario, we underline the fact that the dialogue state representation generated by the Guesser model is used to recover several types of attributes associated with the target object. In this work, we use the same dialogue state representation as used by \newcite{shekhar2019beyond} and only focus on improving the object representations using the imagination component. Indeed, the best imagination model \textsc{GDSE-SL+Imagination} is in line with \textsc{GDSE-SL}, currently the best model in terms of attribute prediction. In particular, even though the dialogue state representation is only indirectly affected by the imagination embeddings (via a dot-product operation to score the candidate objects), we can still see an improvement in terms of F1 for \textit{Location} attributes (L-F1) and similar performance for \textit{Situated} attribute prediction (S-F1). Both can be considered, to some extent, a result of better situated object representations.

\paragraph{Zero-shot gameplay.}
As underlined in Section~\ref{sec:imagination}, the imagination module's main strength is to be able to distill imagination embeddings from perceptual information only, without relying on externally provided category labels. The zero-shot gameplay scenario from \compguesswhat{}\  (Table~\ref{tab:compguesswhat}) sheds some light on the ability of the model to generalize to out-of-distribution examples. In the out-of-domain gameplay scenario where candidate objects belonging to categories never seen before are present, both imagination-based models \textsc{GDSE-SL+Imagination} and \textsc{GDSE-CL+Imagination} outperform the previous best performing system \textsc{DeVries-RL} by $1.2\%$ and $8.26\%$, respectively in terms of OD accuracy (\textsc{OD-Acc}). 
By analyzing their output, we notice that the best imagination model achieves higher accuracy by learning a better gameplay strategy involving half the amount of  location questions generated by \textsc{DeVries-RL} ($39.68\%$ vs $75.84\%$; see Appendix~\ref{appendix:error_analysis} for more details). A further improvement in the near-domain scenario (\textsc{ND-Acc}) confirms the effectiveness of the imagination component to generate category embeddings for objects on-the-fly using only perceptual information. 
\begin{figure}[!b]
\centering
\includegraphics[keepaspectratio, scale=0.54]{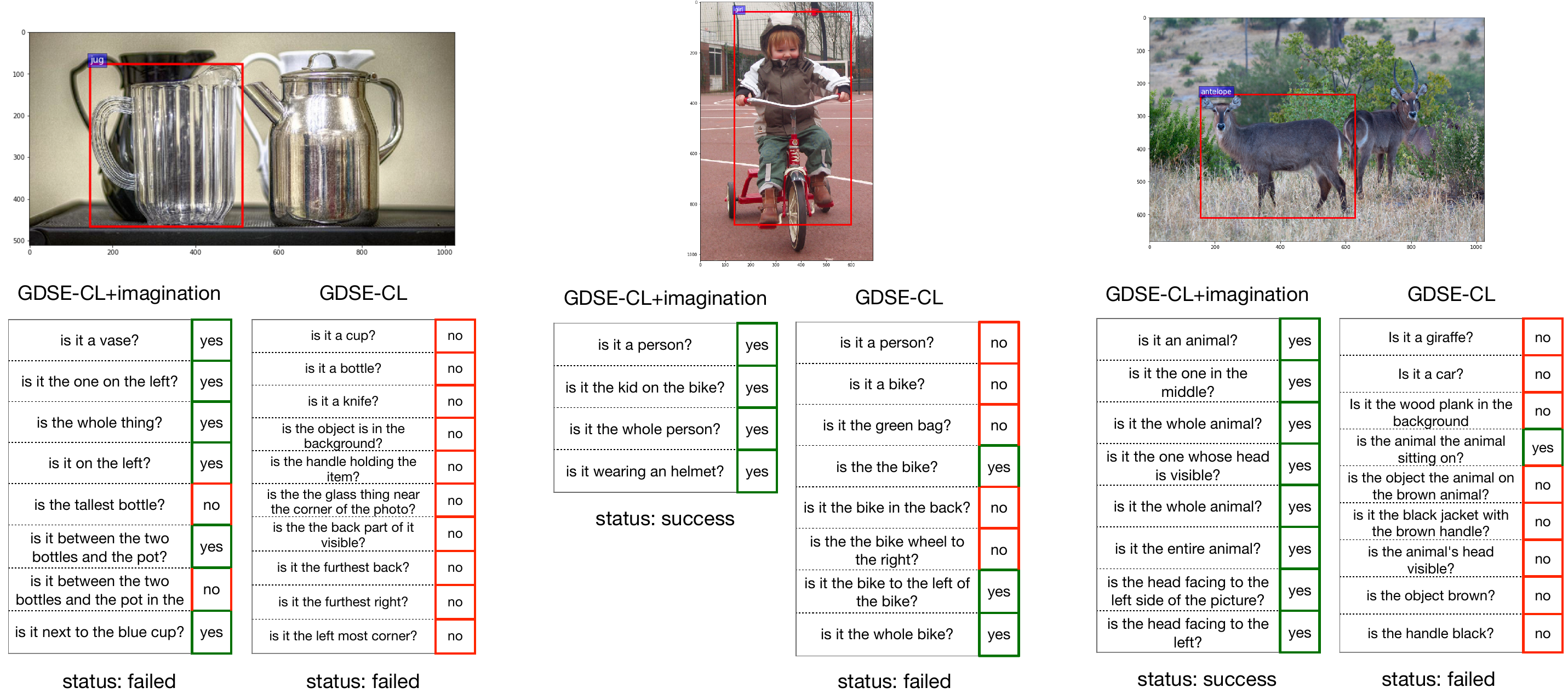}
\caption{Qualitative examples in the zero-shot gameplay scenario: the categories 'girl' and 'antelope' are not present in MSCOCO and therefore cannot be encoded by the GDSE-CL model. On the other hand, the imagination model is able to \textit{distill} imagination embeddings by using the crop features only (for the sake of presentation quality we remove consecutive repeated questions).}
\label{fig:od_test_examples}
\end{figure}

\paragraph{Out-of-domain error analysis.} 
Lastly, we report an error analysis comprising 50 dialogues selected at random from out-of-domain games (for more details refer to Appendix~\ref{appendix:error_analysis}). First, we manually annotated the Oracle answers and partitioned them according to their type using the same question classifier used for the Oracle Task (Section~\ref{sec:oracle_results}). 
$83\%$ of super-category questions (from a total of $80$) were correctly answered by the model and $63.36\%$ color related questions (from a total of $88$) were correctly answered. For instance, as shown in Figure~\ref{fig:od_test_examples}, \textsc{GDSE-CL} is not able to answer correctly the question ``is it a person?'' because it does not have category information for the label ``girl'' but only for the label ``person''. 
On the other hand, \textsc{GDSE-CL+imagination} is able to a) categorize the object as a member of the super-category ``person'', and b) correctly ground the expression ``kid on the bike'' to the target object. The same behavior can be observed when the ``antelope'' is the target object. Antelopes are not part of the MSCOCO classes, and therefore have not been seen by the model during training. First, the model refers to it as ``animal'', hence the Oracle is able to correctly answer the question even though ``antelope'' was never involved in the training. Secondly, we found that the number of $No$ answers for \textsc{GDSE-CL} is considerably higher ($88.06\%$) than \textsc{GDSE-CL+imagination} ($51.02\%$), validating our hypothesis that the Oracle does not know how to deal with unseen instances. Finally, in the imagination dialogue of the first example, 
even though the generated question/answers were probably referring to the correct object, the Guesser model is eventually unable to guess correctly. More work is required to better fuse the language modality and the object representations to improve its performance.

\section{Related Work}
\label{sec:related}
Concerning unsupervised learning of concept representations,
\newcite{bruni2014multimodal} first learn modality-specific representations and then fuse them into a unified representation for each concept.
However, they rely on hand-crafted bags of visual features, making the approach laborious to extend to new domains and games.
\newcite{kiela2018learning} cope with this issue by relying on CNN models to extract latent features from images for instances of specific objects.
\newcite{lazaridou2015combining} use a margin loss but in the context of maximizing the similarity between the visual representation of a noun phrase and its corresponding text representation. 
Similarly,  \newcite{collell2017imagined} learn a mapping between the ResNet features and the word embeddings of a concept.
As discussed in Section~\ref{sec:background}, unlike our imagination embeddings, these purely-perceptual representations are neither category-aware nor context-aware.
\newcite{silberer2016visually} present a multi-modal model that uses a denoising auto-encoder framework. 
Unlike us, they do not use perceptual information as input but rely on an attribute-based representation derived from an additional attribute predictor. 
However, they do use a reconstruction loss (cross-entropy loss for attribute prediction) and an auxiliary category loss during training. 
Their training scheme is more complex as they first separately train the AE for each modality and then fuse them, which we avoid by adopting a single end-to-end architecture. 
\newcite{ebert-pavlick-2019-using} used VAEs to learn grounded representations for lexical concepts. 
However, as discussed in Section~\ref{sec:imagination}, VAEs are not as well suited as RAEs to representation learning for our imagination module. In the context of guessing games, all the previous approaches rely on categories embeddings~\cite{de2017guesswhat,shekhar2019beyond,strub2017end,zhuang2018parallel,shukla2019should} (see Section~\ref{sec:background}).
Our imagination component can be flexibly integrated in any of them by replacing the category embeddings with imagination embeddings.

\section{Conclusions}

We argued that  existing models for learning grounded conceptual representations fail to learn compositional and generalizable multi-modal representations, relying instead on the use of category labels for every object in the scene both at training and inference time~\cite{de2017guesswhat}.   To address this, we introduced a novel “imagination” module based on Regularized Auto-Encoders, that learns a context-aware and category-aware latent embedding for every object directly from its image crop, without using category labels. We showed state-of-the-art performance in the CompGuessWhat?! zero-shot scenario~\cite{suglia2020compguesswhat}, outperforming current models by 8.26\% in gameplay accuracy while performing comparably on the other tasks to models which use category labels at training time. The imagination-based model also shows improvements of 2.08\% and 12.86\% in Oracle and Guesser accuracy.  Finally, we conducted an extensive error analysis and showed that imagination embeddings help to reason about object visual properties and attributes.
For future work, we plan to 1) integrate category labels at training time  in a more principled way following advances in semi-supervised learning~\cite{kingma2014semi}; 2) improve the multi-task learning procedure presented in \cite{shekhar2019beyond} to optimize at the same time multiple tasks of different complexities.

\bibliographystyle{coling}
\bibliography{coling2020}

\clearpage

\appendix
\section{Appendix}

\subsection{Model details}\label{appendix:model_details}

As described in Section 3 of the main paper, we extend both the Oracle and Guesser model with an imagination component. For both roles, we keep the same model structure for the imagination component. In this paper we implement $\En$ as a 2-layer feed-forward neural network with ReLU~\cite{dahl2013improving} activation function. We acknowledge that many other implementations are possible in this case and we leave more complex designs for future work. Given the latent code $\emb_i$ generated by the function $\En$, we use a decoder $\D$ to generate the reconstructed perceptual input (\emph{imagined}) of the object $o_i$, $\D(\emb_i) = \tilde{\objemb}_{i}$. As common practice, we define the decoder $\D$ as symmetric to the architecture of the encoder $\En$. For the category embeddings size $d_c$, as in \cite{shekhar2019beyond}, we use $256$ and $512$ for the Oracle and Guesser respectively. For the imagination component, we run a grid search involving several parameters for the latent code $\emb$ such as $(16, 32, 64, 128, 256, 512)$. For both roles, we choose $512$ because it was the value that lead to the highest accuracy on the validation set. We also experimented with several values for the coefficient $\alpha$ of the regularization term \lossreg: (1e-3, 1e-5, 1e-6, 1e-7). For the Oracle the best value resulted to be $1e-7$, while $1e-5$ for the Guesser. When training the imagination component with the object category loss, due to the class imbalance, we apply loss weighting. We compute the class weights using the method reported in \cite{king2001logistic}. For the margin value $\eta$ we opted for $1.0$ after experimenting with a less effective dynamic margin that would change depending on the distance between the concepts in the WordNet hierarchy.

\subsection{Training details}\label{appendix:training_details}

For both roles, we train the models using the Adam optimizer~\cite{kingma2014adam}. For the Oracle and Guesser training we use $0.0001$ as learning rate. In both cases, we use the original \guesswhat{}\ validation set to select the best model that is used in the evaluation on the test set. As described in \cite{shekhar2019beyond}, we use a modulo-$n$ training procedure to jointly optimize both the Guesser and Questioner. In our experimental evaluation we run a grid search of several values of $n$ such as $3, 5, 7$. We selected $5$ as the best performing value on the validation set. For a fair comparison with all the GDSE model variants trained with Supervised Learning and Collaborative Learning, we made the same architectural choices and hyperparameters values. Please refer to the original codebase implementation available on GitHub~\footnote{\url{https://github.com/shekharRavi/Beyond-Task-Success-NAACL2019}}. Another point of difference is in the Collaborative Learning fine-tuning phase for the Guesser model. During this phase, only the Questioner and Guesser models are fine-tuned whereas the Oracle model is fixed~\cite{shekhar2019beyond} therefore, we decided to use the best performing Oracle so that the Guesser model is not negatively affected by a less performing Oracle and also to be comparable with the original implementation. 

\subsection{Error analysis}\label{appendix:error_analysis}

In order to provide a more fine-grained evaluation of the generated dialogues, we adapt the quality evaluation script presented by \newcite{suglia2020compguesswhat} and extend it with additional metrics. First of all, it relies on a rule-based question classifier that classifies a given question in one of seven classes: 1) super-category (e.g., ``person", ``utensil", etc.), 2) inanimate object (e.g., ``car", ``oven", etc.), 3) animate object (e.g., ``dog", ``cat", etc.), 3) ``color", 4) ``size", 5) ``texture", 6) ``shape" and ``location". The question classifier is useful to evaluate the dialogue strategy learned by the models. In particular, we look at two types of turn transitions: 1) super-category $\rightarrow$ object/attr, it measures how many times a question with an affirmative answer from the Oracle related to a super-category is followed by either an object or attribute question (where ``attribute" represents the set $\{$color, size, texture, shape and location$\}$; 2) object $\rightarrow$ attr, it measures how many times a question with an affirmative answer from the Oracle related to an object is followed by either an object or attribute question. We compute the \textit{lexical diversity} as the type/token ratio among all games, \textit{question diversity} and the percentage of games with repeated questions. We also evaluate the percentage of dialogue turns involving location questions. Table \ref{tab:nd_test_gameplay_quality} and \ref{tab:od_test_gameplay_quality} show the results of these analysis for the models \texttt{GDSE-CL} and \texttt{GDSE-CL+imagination} analyzed in this paper. 

Using the above-mentioned question classifier, we completed an error analysis trying to understand the quality of the generated gameplay in a zero-shot scenario from the point of view of the answers prediction performance and the guesser accuracy. In particular, we randomly sampled a pool of 50 reference games from the out-of-domain zero-shot scenario and we manually annotated whether a given answer generated by the Oracle model was correct or not. Table~\ref{tab:od_oracle_error_analysis} shows the results of the manual annotation step. The model confirms high performance in answering questions about super-category information demonstrating that it is able to correctly categories objects in macro-categories even though is has not seen them before. 

\subsubsection{Zero-shot gameplay quality}\label{appendix:zero_shot_quality}

\begin{table}[h]
\centering
\resizebox{\textwidth}{!}{%
\begin{tabular}{@{}llllllllll@{}}
\toprule
\textbf{Model} & \textbf{\begin{tabular}[c]{@{}l@{}}Lexical\\ diversity\end{tabular}} & \textbf{\begin{tabular}[c]{@{}l@{}}Question\\ diversity\end{tabular}} & \textbf{\begin{tabular}[c]{@{}l@{}}\% games \\ repeated \\ questions\end{tabular}} & \textbf{\begin{tabular}[c]{@{}l@{}}Super-cat -\textgreater \\ obj/attr\end{tabular}} & \textbf{\begin{tabular}[c]{@{}l@{}}Object -\textgreater \\ attribute\end{tabular}} & \textbf{\begin{tabular}[c]{@{}l@{}}\% turns\\ location \\ questions\end{tabular}} & \textbf{Vocab. size} & \textbf{Accuracy} \\ \midrule
DeVries-RL & 0.13	& 1.77 &	99.48	& 97.39	& 98.70	& 78.07	& 702.00	& 43.92\% \\
GDSE-CL               & 0.17              & 13.74              & 66.75                          & 93.62                            & 66.27                          & 31.23                               & 1260    & 43.42\%                  \\
GDSE-CL + Imagination & 0.10              & 8.56               & 91.80                          & 93.15                            & 60.72                          & 39.90                               & 808      & 46.70\%                \\ \bottomrule
\end{tabular}
}
\caption{Comparison between the quality of gameplay in the near-domain zero-shot scenario between GDSE-CL and GDSE-CL with imagination. Number of total turns $10$.}
\label{tab:nd_test_gameplay_quality}
\end{table}

\begin{table}[h]
\centering
\resizebox{\textwidth}{!}{%
\begin{tabular}{@{}llllllllll@{}}
\toprule
\textbf{Model} & \textbf{\begin{tabular}[c]{@{}l@{}}Lexical\\ diversity\end{tabular}} & \textbf{\begin{tabular}[c]{@{}l@{}}Question\\ diversity\end{tabular}} & \textbf{\begin{tabular}[c]{@{}l@{}}\% games \\ repeated \\ questions\end{tabular}} & \textbf{\begin{tabular}[c]{@{}l@{}}Super-cat -\textgreater \\ obj/attr\end{tabular}} & \textbf{\begin{tabular}[c]{@{}l@{}}Object -\textgreater \\ attribute\end{tabular}} & \textbf{\begin{tabular}[c]{@{}l@{}}\% turns\\ location \\ questions\end{tabular}} & \textbf{Vocab. size} & \textbf{Accuracy} \\ \midrule
DeVries-RL & 0.24	& 2.96 &	98.49	& 91.26	& 98.57	& 75.84	& 1275	& 38.73\% \\
GDSE-CL               & 0.14              & 7.86               & 66.32                          & 91.67                            & 72.33                          & 26.03                               & 1002    & 29.83\%    &                      \\
GDSE-CL + Imagination & 0.10              & 8.57               & 89.19                          & 94.82                            & 58.51                          & 39.68                               & 814      & 46.93\%                  \\ \bottomrule
\end{tabular}
}
\caption{Comparison between the quality of gameplay in out-of-domain zero-shot scenario between GDSE-CL and GDSE-CL with imagination. Number of total turns $10$.}
\label{tab:od_test_gameplay_quality}
\end{table}

\begin{table}[!htb]
    \begin{minipage}{.5\linewidth}
     \centering
      \begin{tabular}{@{}lll@{}}
        \toprule
        \textbf{Question type}    & \textbf{Accuracy} & \textbf{Count} \\ \midrule
        Inanimate object & 65.48\%  & 168   \\
        Animate object   & 53.33\%  & 15    \\
        Super category   & 83.33\%  & 60    \\
        Location         & 78.86\%  & 175   \\
        Size             & 100.00\% & 1     \\
        Color            & 58.33\%  & 24    \\
        Parts            & 100.00\% & 2     \\ \bottomrule
        \end{tabular}
    \end{minipage}%
    \begin{minipage}{.5\linewidth}
      \centering
        \begin{tabular}{@{}lll@{}}
        \toprule
        \textbf{Question type}    & \textbf{Accuracy} & \textbf{Count} \\ \midrule
            Inanimate object & 81.71\%  & 164   \\
            Animate object   & 70.00\%  & 10    \\
            Super category   & 67.61\%  & 71    \\
            Location         & 72.97\%  & 148   \\
            Size             & 100\% &  1    \\
            Color            & 63.64\%  &  88   \\
            Parts            & 71.43\% &7      \\ \bottomrule
        \end{tabular}
    \end{minipage} 
    \caption{Error analysis results completed on the Out-of-domain zero-shot scenario for the model \texttt{GDSE-CL+Imagination} (on the left) and \texttt{GDSE-CL} (on the right).}
    \label{tab:od_oracle_error_analysis}
\end{table}

\end{document}